\documentclass[10pt]{article}

\usepackage[margin=0.68in]{geometry}
\usepackage{booktabs}
\usepackage{array}
\usepackage{graphicx}
\usepackage{amsmath,amssymb}
\usepackage{microtype}
\usepackage{url}
\usepackage[hidelinks]{hyperref}
\usepackage{caption}
\usepackage{subcaption}
\usepackage{float}
\usepackage{enumitem}
\usepackage{xcolor}
\usepackage{tikz}
\usetikzlibrary{arrows.meta,positioning,fit,calc,shapes.misc}
\setlist{nosep,leftmargin=*}
\definecolor{navy}{HTML}{173B6C}
\definecolor{teal}{HTML}{0F6B73}
\definecolor{violet}{HTML}{5A3E8A}
\definecolor{orange}{HTML}{C96B12}
\definecolor{redsoft}{HTML}{B33A3A}
\definecolor{lightblue}{HTML}{EEF4FB}
\definecolor{lightteal}{HTML}{ECF8F7}
\definecolor{lightviolet}{HTML}{F4F0FA}
\definecolor{lightorange}{HTML}{FFF4E8}
\definecolor{lightred}{HTML}{FCEEEE}
\definecolor{graybox}{HTML}{F5F6F7}

\newcommand{\tightsection}[1]{\vspace{0.1em}\section{#1}\vspace{-0.25em}}

\begin{document}

\begin{center}
{\LARGE\bfseries
When Minute-Resolution Monitoring Meets Session-Level Injury Labels:\\[-0.1em]
Landmark-Based Discrimination in Elite Women's Football\par}
\end{center}

\vspace{0.35em}

\begin{figure}[h!]
\centering
\includegraphics[width=0.99\linewidth]{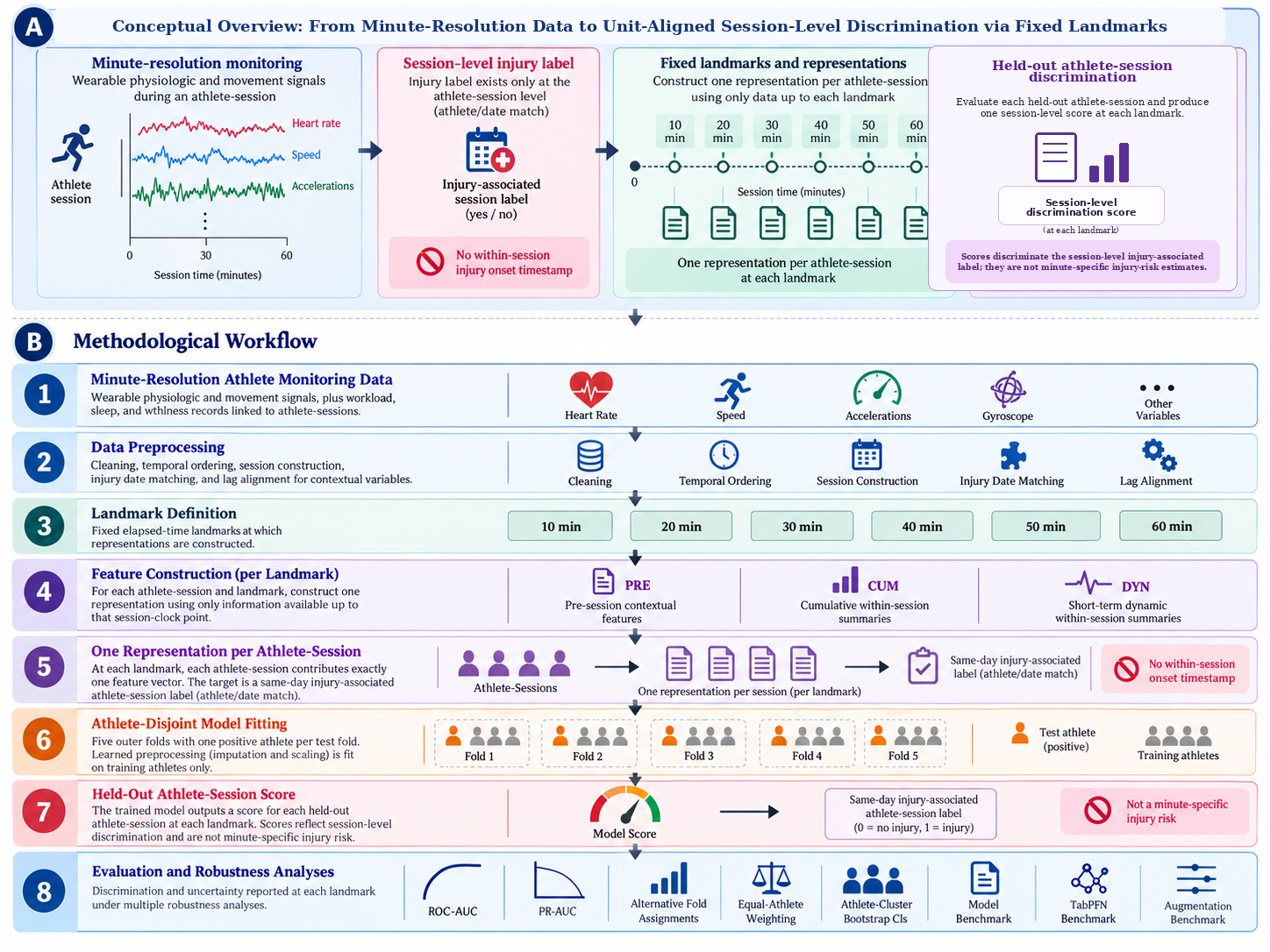}
\caption{Conceptual overview and methodological workflow. Minute-resolution monitoring is paired with a same-day athlete-session injury label whose within-session onset time is unknown. The proposed analysis therefore constructs one representation per athlete-session at fixed landmarks, preserves athlete-disjoint evaluation, and treats model benchmarking and synthetic augmentation as downstream analyses rather than as sources of finer-grained outcome supervision.}
\label{fig:overview}
\end{figure}

\vspace{-0.25em}
\begin{center}
{\large Evangelos Chatzidimitriou \qquad Konstantinos Tserpes\par}
\vspace{0.12em}
{\small National Technical University of Athens, Athens, Greece\par}
\end{center}

\vspace{0.45em}
\begin{abstract}
\textbf{Introduction.} Minute-resolution athlete monitoring is increasingly common, but injury annotation may exist only at the athlete-session level and may omit the within-session onset time. Replicating a positive session label across every recorded minute would therefore create unsupported minute-level supervision. We study how to use high-resolution within-session information without making that error.

\textbf{Methods.} Using 2020 SoccerMon data from elite women's football, we construct exactly one representation per athlete-session at fixed 10, 20, 30, 40, 50, and 60 min landmarks using only information observed by that session-clock point. The target remains a same-day injury-associated athlete-session indicator. The reconstructed resource contains 3,743 athlete-sessions from 48 athletes; all 22 positive sessions arise from five athletes in Team A, yielding a 2,259-session, 27-athlete modelling cohort. Evaluation is athlete-disjoint. We compare pre-session (PRE), cumulative (CUM), and dynamic (DYN) representations; logistic regression, Random Forest, XGBoost, and TabPFN; and training-only NONE, SMOTE, and CTGAN conditions. Uncertainty and robustness are examined with paired athlete-cluster bootstrap, a fixed 60-min common cohort, 100 alternative negative-athlete fold allocations, leave-one-positive-athlete-out sensitivity, and equal-athlete weighting.

\textbf{Results.} Primary CUM+DYN logistic-regression discrimination is non-monotonic (ROC-AUC 0.499, 0.557, 0.607, 0.428, 0.402, 0.367 across 10--60 min). The same qualitative pattern persists on the fixed 2,104-session common cohort and across alternative fold allocations. TabPFN reaches ROC-AUC 0.692 at 30 min and improves later-landmark discrimination relative to logistic regression, but does not consistently outperform Random Forest. Synthetic augmentation is not universally beneficial: Random Forest shows positive condition-specific effects (e.g., 30-min SMOTE and 40-min CTGAN), whereas logistic regression shows little systematic benefit and several negative CTGAN contrasts. CTGAN shows no near-duplicate memorization signal but substantial distributional drift.

\textbf{Conclusion.} The contribution is not minute-specific injury prediction. It is a unit-aligned framework for extracting temporal information from minute-resolution monitoring when supervision is only session-level. Within this cohort, results depend materially on landmark, representation, learner, and estimand, while inferential precision remains limited by only five injury-positive athletes.
\end{abstract}

\noindent\textbf{Keywords:} athlete monitoring; women's football; weak supervision; landmark analysis; injury-associated sessions; tabular foundation models; synthetic augmentation.

\tightsection{Introduction}

Sports monitoring systems can record physiology and movement at minute or sub-minute resolution, while injury information is often recorded much more coarsely. This asymmetry matters. High temporal resolution in the predictors does not imply high temporal resolution in the target. If an injury report identifies an athlete and calendar date but not the within-session onset time, a positive session cannot support a claim that every minute was an ``injury minute,'' nor a minute-specific probability trajectory or operational alert.

This is a label-resolution mismatch. It is easy to obscure statistically because the monitoring table may contain hundreds of thousands of minute rows, yet the number of independently informative positive athletes can remain very small. Reviews of sports-injury prediction repeatedly identify small effective samples, heterogeneous outcome definitions, imbalance, and weak validation as major threats to inference \cite{vaneetvelde2021,majumdar2022,bullock2022,aslani2026}. Our focus is a specific additional failure mode: treating dense predictor rows as if they carried equally dense outcome supervision.

We address that problem with a fixed-landmark, one-representation-per-athlete-session formulation. At six elapsed-time landmarks, a session is summarized using only information observed up to that session-clock point, while the outcome remains the original session-level injury-associated label. The formulation is related to landmarking as an organizing principle for longitudinal prediction \cite{vanhouwelingen2007}, but differs from classical event-history landmarking because no time-to-event target is available. It is also distinct from multiple-instance learning because we do not attempt to localize a latent causal minute within a positive session \cite{dietterich1997}.

The paper makes four contributions. First, it defines a unit-aligned framework for high-frequency monitoring with coarse session-level injury supervision. Second, it evaluates how discrimination changes with landmark and feature representation under athlete-disjoint validation and cluster-aware uncertainty. Third, it compares conventional learners with TabPFN, a tabular foundation model, under the same held-out athlete structure. Fourth, it tests whether training-only SMOTE or CTGAN augmentation changes held-out discrimination, while explicitly auditing synthetic-data fidelity and evaluating sensitivity to the very small number of positive athletes. The goal is not a model-selection leaderboard; it is to characterize what this supervision regime can and cannot support.

\tightsection{Related Work}

\textbf{Athlete monitoring and injury modelling.} Training-load, wellness, sleep, physiological, and locomotor variables are widely used to characterize athlete state \cite{halson2014,saw2016,bartlett2017,bourdon2017}. Football injury modelling spans session-level risk-factor analysis, longitudinal forecasting, and diverse machine-learning targets \cite{bartels2024,catterall2026,majumdar2022,aslani2026}. Those tasks are related but not directly comparable when their outcome definitions, temporal units, or validation schemes differ.

\textbf{Temporal granularity.} Wearable conclusions can depend on observation epoch even when the outcome itself is not injury-related \cite{baptista2024}. In the present setting, the more fundamental issue is that the predictor clock is fine-grained while the injury label is coarse. The analysis unit must therefore be chosen to match what the label actually identifies.

\textbf{Imbalanced and small tabular learning.} SMOTE creates synthetic minority examples by interpolation \cite{chawla2002}; CTGAN models mixed-type tabular distributions with a conditional generative adversarial network \cite{xu2019}. TabPFN uses a pretrained transformer prior for small tabular classification problems \cite{hollmann2025}. None of these methods creates new independent athletes or new adjudicated injury events; in this study they are evaluated only as modelling interventions inside the training folds.

\tightsection{Methods}

\subsection{Problem Formulation}
Let athlete $a$ participate in session $s$ with session-level label $Y_{a,s}\in\{0,1\}$. For landmark $\ell\in\{10,20,30,40,50,60\}$ min, define
\[
z^{(\ell)}_{a,s}=g\!\left(X_{a,s,1:\ell},C^{\mathrm{pre}}_{a,s}\right),
\qquad
\hat p^{(\ell)}_{a,s}=f_{\ell}\!\left(z^{(\ell)}_{a,s}\right).
\]
$X_{a,s,1:\ell}$ contains within-session monitoring observed no later than landmark $\ell$, while $C^{\mathrm{pre}}_{a,s}$ contains contextual variables assigned under the preprocessing rules below. Each athlete-session contributes at most one observation to a given landmark. $\hat p^{(\ell)}_{a,s}$ is interpreted only as a discrimination score for the same-day injury-associated session label. It is not an estimate of injury onset, not a minute-specific prospective injury probability, and not an operational alert.

\subsection{Dataset and Cohort Accounting}
We use the 2020 season of the SoccerMon dataset \cite{midoglu2024}, collected from teams in elite Norwegian women's football. Minute-level records are organized into team-by-calendar-day exposure units. Within a team-day, gaps greater than 30 min define separate temporal blocks; only team-days with one retained block are used to avoid ambiguous attribution, and a minimum 30-min exposure duration is required.

The reconstructed resource contains 380,193 minute rows, 251 retained team-date exposure units, 3,743 athlete-sessions, and 48 athletes. Injury labels are derived by athlete/date matching to available athlete-submitted injury reports; exact within-session onset is unavailable. Twenty-two injury-associated athlete-sessions arise from five athletes. All positives occur in Team A, which contains 2,259 athlete-sessions from 27 athletes; Team B contains 1,484 sessions and no positives, so it cannot serve as an injury-positive external discrimination cohort. Table~\ref{tab:cohort} summarizes the cohort.

\begin{table}[t]
\centering
\caption{Cohort accounting after preprocessing.}
\label{tab:cohort}
\begin{tabular}{lr}
\toprule
Quantity & Count\\
\midrule
Minute-level observations & 380,193\\
Retained team-date exposure units & 251\\
Athlete-sessions & 3,743\\
Athletes & 48\\
Injury-associated athlete-sessions & 22\\
Athletes contributing positive sessions & 5\\
Team A modelling sessions / athletes & 2,259 / 27\\
Team B sessions / positives & 1,484 / 0\\
\bottomrule
\end{tabular}
\end{table}

At 10, 20, 30, 40, 50, and 60 min, respectively, 2,258, 2,258, 2,257, 2,250, 2,218, and 2,104 Team-A sessions remain evaluable; all 22 positives remain represented at every landmark. Denominator reduction is therefore driven by shorter negative sessions.

\subsection{Feature Construction and Representations}
PRE contains 14 contextual predictors: seven workload variables, two sleep variables, and five wellness variables. Raw daily dates are parsed explicitly as day-month-year. Workload and wellness variables are lagged by one calendar day. Sleep duration and quality are matched to the session date and interpreted as descriptors of the preceding night; the source data do not provide a verified pre-session submission timestamp for these same-day sleep fields. PRE is therefore treated as a contextual sensitivity family rather than a deployment-ready real-time feature set.

Within-session features are derived from ten sensor signals covering speed, heart rate, acceleration, acceleration impulse, and accelerometer/gyroscope variability. CUM contains 30 expanding summaries from session start through the current minute. DYN contains 33 short-horizon dynamic summaries, including first differences and trailing five-observation rolling summaries. All temporal operations are grouped by athlete-session and use no later-minute information.

We evaluate PRE (14), CUM (30), DYN (33), PRE+CUM (44), PRE+DYN (47), CUM+DYN (63), and ALL (77). CUM+DYN is the primary representation because it uses only sensor-derived within-session features and avoids dependence on the more incomplete and timing-ambiguous contextual variables.

\subsection{Athlete-Disjoint Evaluation and Uncertainty}
Evaluation uses deterministic five-fold athlete-disjoint outer validation. The five athletes with at least one positive session are assigned one per test fold; negative-only athletes are distributed across folds. No athlete appears in both training and test data within a fold. Imputation, scaling, synthetic augmentation, and model fitting are trained only on outer-training athletes. Held-out fold predictions are pooled so that each evaluable athlete-session contributes exactly one out-of-athlete prediction at each landmark.

The primary classifier is L2-regularized logistic regression ($C=1.0$, balanced class weights, \texttt{liblinear}, maximum 2,000 iterations, seed 42). Performance is summarized with ROC-AUC and PR-AUC; because positive prevalence is approximately 1\%, PR-AUC is interpreted relative to the prevalence baseline \cite{saito2015}. Repeated sessions from the same athlete are dependent, so 95\% intervals are obtained from 1,000 athlete-cluster bootstrap resamples of pooled out-of-fold predictions. Resamples containing only one outcome class are discarded. Paired model and augmentation contrasts reuse the same athlete bootstrap draw within each comparison. These intervals quantify cluster-level sampling uncertainty conditional on the fitted cross-validation procedure; they do not capture every source of model-training uncertainty.

\subsection{Robustness and Estimand Sensitivity}
Three analyses probe the primary temporal pattern. First, every landmark is re-evaluated on the same 2,104 sessions observable through 60 min. Second, the one-positive-athlete-per-test-fold rule is held fixed while negative athletes are reassigned across folds in 100 reproducible random allocations. Third, an equal-athlete-weight estimand assigns each athlete the same total weight at a landmark, rather than weighting athletes in proportion to their number of evaluable sessions. These are sensitivity analyses, not replacements for the primary session-pooled estimand.

\subsection{Model-Family Benchmark}
Logistic regression, Random Forest, XGBoost, and TabPFN are evaluated on the same CUM+DYN landmark datasets and athlete-disjoint folds. Random Forest uses 500 trees, unrestricted depth, minimum leaf size 2, balanced class weights, and seed 42. XGBoost uses 300 trees, depth 3, learning rate 0.03, row and feature subsampling 0.8, and fold-specific positive-class weighting. Hyperparameters are fixed rather than selected by additional tuning. TabPFN is evaluated under the same outer test assignments and features. The frozen rerun reproduced TabPFN ROC-AUC and PR-AUC values to maximum absolute differences below $4\times10^{-7}$ and $5\times10^{-7}$, respectively.

\subsection{Training-Only Synthetic Augmentation}
Augmentation is evaluated only for logistic regression and Random Forest to avoid a large model-by-generator search. For each outer fold and landmark, NONE, SMOTE, and CTGAN training conditions are compared. Augmentation ratios are 2.5\%, 5\%, and 10\% of the real training-set size, with seeds 11, 22, 33, 44, and 55. SMOTE operates after training-fold median imputation and standardization; generated samples are returned to the original feature scale before being appended to the real training rows. CTGAN is fitted only to positive training rows. Held-out athletes are never augmented.

The full design contains 930 normalized training units and 1,860 model fits. Augmentation effects are computed as the mean across the five augmentation seeds relative to the matched NONE baseline, then evaluated with 1,000 shared-draw athlete-cluster bootstrap resamples. Because 72 augmentation contrasts are exploratory and no multiplicity-adjusted confirmatory procedure was prespecified, we describe intervals as excluding or including zero rather than labeling isolated contrasts as statistically significant.

CTGAN quality is audited separately at the 30-min fold-1 seed-11 condition using 200 synthetic positives and 18 real positive training sessions. For the nearest-neighbour audit, all 63 features are standardized using a scaler fitted only on the real positive training rows; Euclidean distances are therefore reported in that real-positive standardized feature space. No exact or near duplicate is observed at distance thresholds below $10^{-6}$ or 0.1. However, the median synthetic-to-nearest-real distance is 10.89 versus 7.30 for real-to-other-real distance (ratio 1.49), median absolute standardized mean difference is 0.789, 49/63 features exceed 0.5, and median absolute correlation change is 0.218. We therefore retain CTGAN as a prespecified augmentation arm but do not claim high-fidelity recovery of the positive distribution.

\subsection{Positive-Athlete and Weighting Sensitivity for Augmentation}
Leave-one-positive-athlete-out summaries assess whether augmentation direction is driven by a single positive athlete. We also recompute augmentation effects under equal-athlete weighting. Across the 72 contrasts, primary and equal-athlete estimands agree in ROC and PR direction in 76.4\% of comparisons, demonstrating that roughly one quarter of effect directions depend on the weighting estimand.

\tightsection{Results}

\subsection{Primary Landmark Pattern and Robustness}
Primary CUM+DYN logistic-regression ROC-AUC is 0.499, 0.557, 0.607, 0.428, 0.402, and 0.367 across 10--60 min; corresponding PR-AUC is 0.0107, 0.0150, 0.0129, 0.0088, 0.0079, and 0.0080. Athlete-cluster intervals are wide, and the largest point estimate at 30 min should not be interpreted as an optimal intervention time.

The same temporal shape persists when all landmarks are restricted to the identical 2,104-session 60-min cohort: ROC-AUC is 0.489, 0.558, 0.597, 0.413, 0.386, and 0.367. Across 100 alternative negative-athlete fold allocations, median ROC-AUC is 0.498, 0.579, 0.609, 0.419, 0.409, and 0.351. Thus, the non-monotonic pattern is not explained solely by changing landmark denominators or one arbitrary assignment of negative athletes.

Equal-athlete weighting changes the magnitude substantially: weighted ROC-AUC is 0.606, 0.746, 0.685, 0.618, 0.446, and 0.423. This does not represent a better model; it answers a different population-weighting question and shows that session-pooled discrimination should not be interpreted as if every athlete contributed equal total influence.

\subsection{Feature-Family Ablation}
\begin{figure}[H]
\centering
\includegraphics[width=0.94\linewidth]{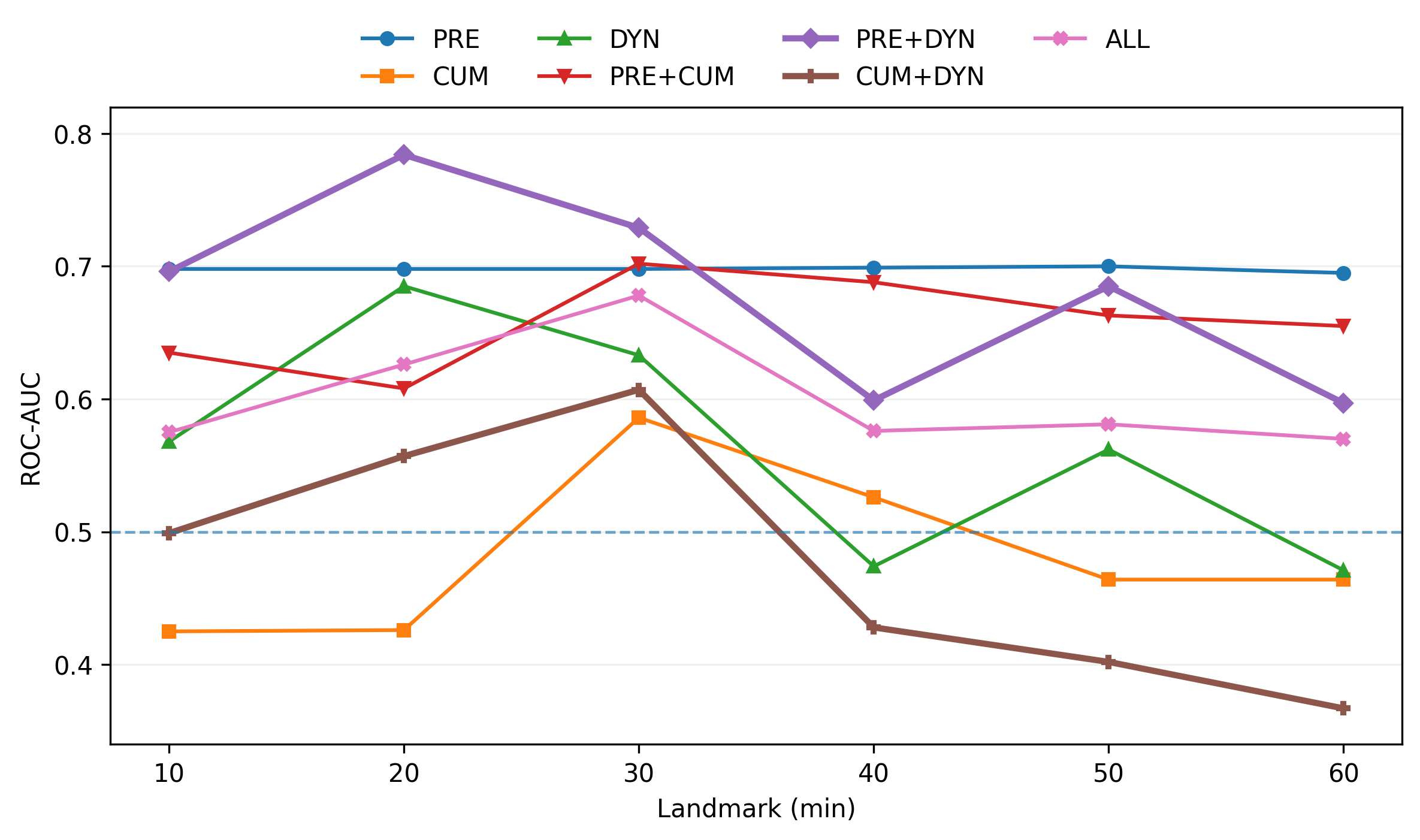}
\caption{Feature-family ROC-AUC trajectories under the fixed logistic-regression evaluation protocol. Relative ordering changes across landmarks; no representation is interpreted as uniformly superior. PRE-containing analyses are contextual sensitivities because of missingness and timing limitations.}
\label{fig:ablation}
\end{figure}

PRE-containing representations yield larger point estimates at several landmarks (Fig.~\ref{fig:ablation}). PRE alone remains near 0.70 ROC-AUC across landmarks, while PRE+DYN reaches 0.784 at 20 min. However, PRE+DYN paired contrasts against primary CUM+DYN include zero at every landmark for both ROC-AUC and PR-AUC. In addition, PRE carries substantial missingness and incompletely verified same-day sleep timing. These results therefore indicate contextual separability rather than deployable pre-session prediction.

\subsection{Model-Family Benchmark}
\begin{figure}[H]
\centering
\begin{subfigure}[t]{0.49\linewidth}
\centering
\includegraphics[width=\linewidth]{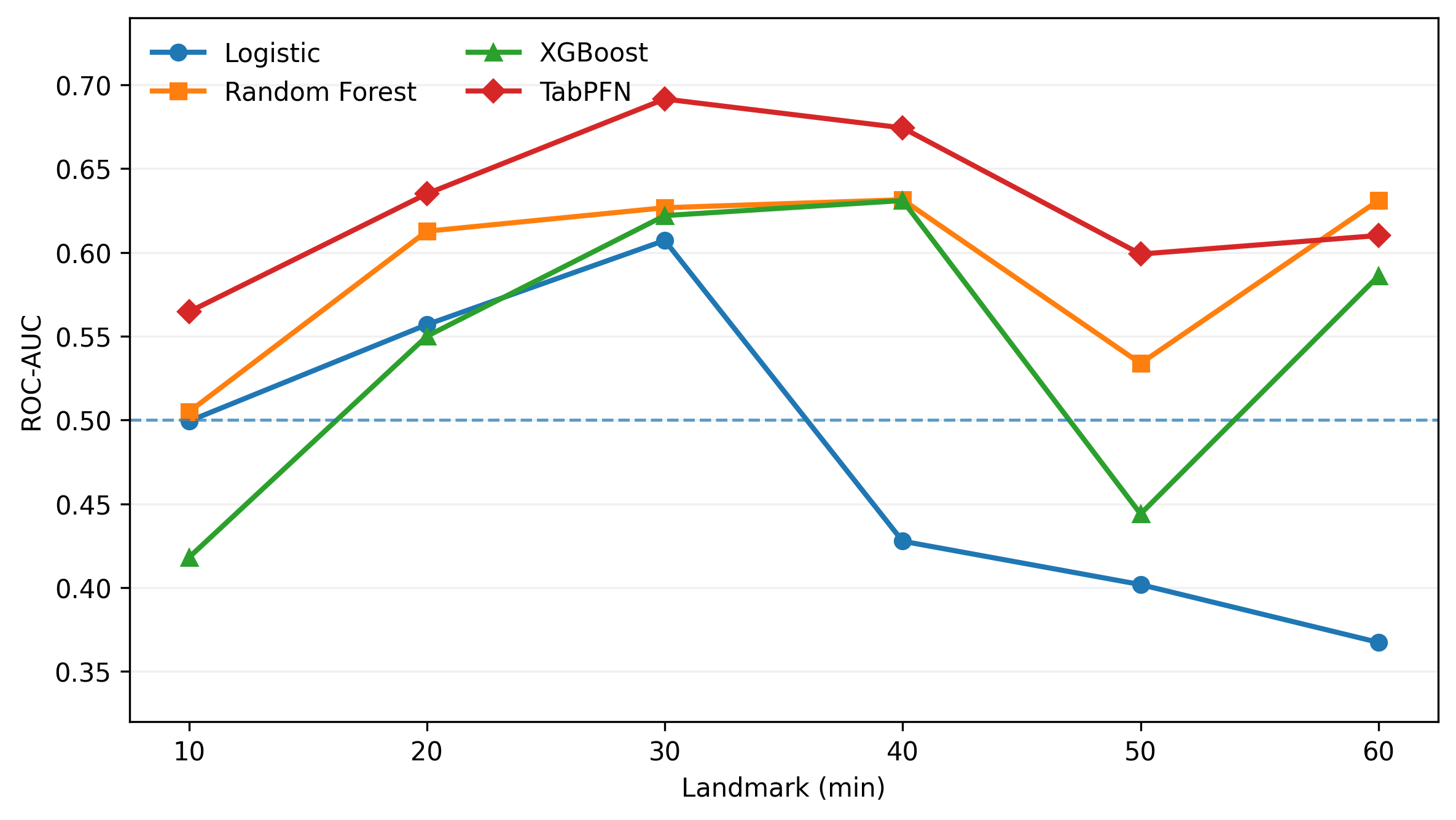}
\caption{ROC-AUC}
\end{subfigure}\hfill
\begin{subfigure}[t]{0.49\linewidth}
\centering
\includegraphics[width=\linewidth]{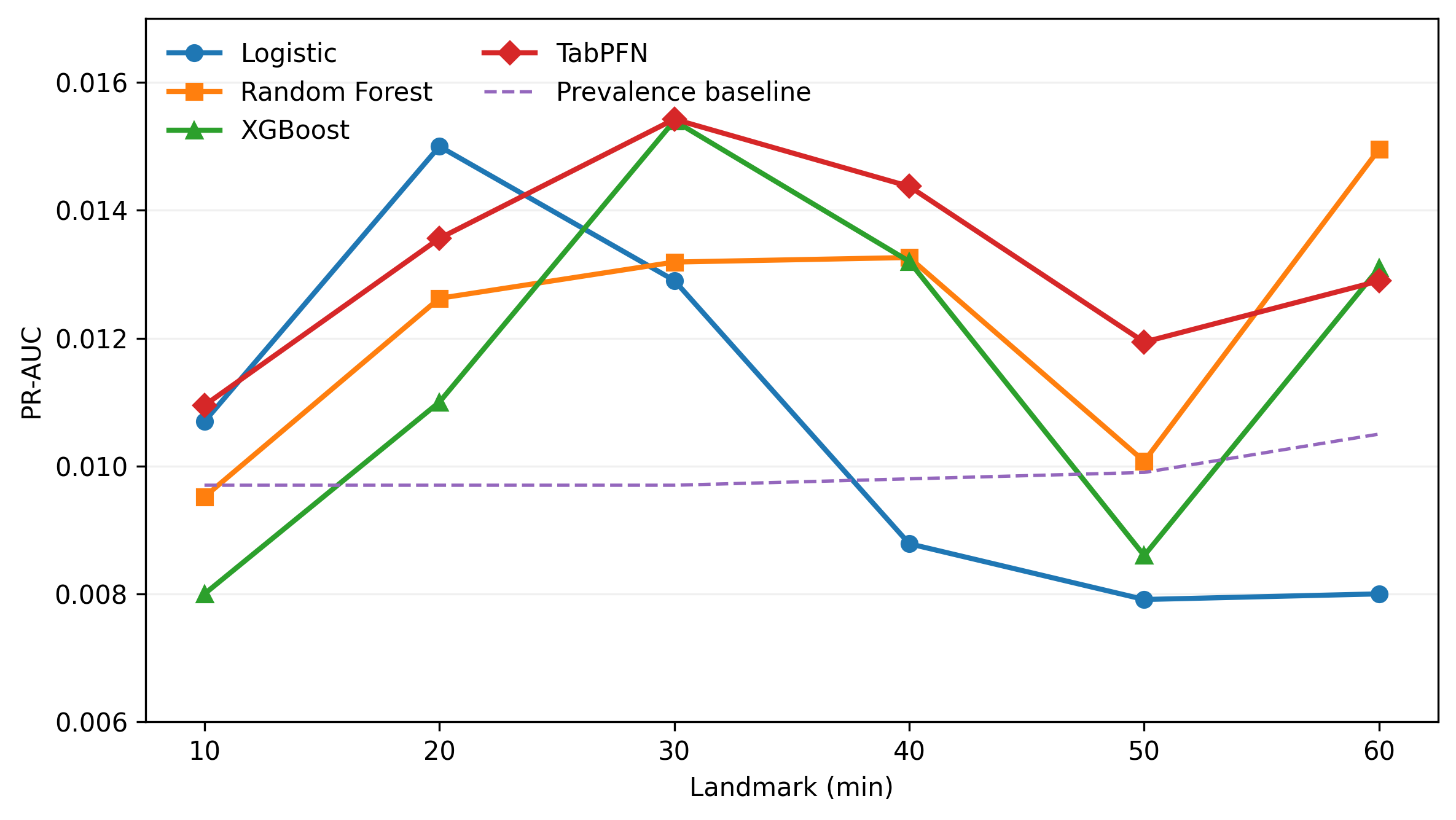}
\caption{PR-AUC}
\end{subfigure}
\caption{Matched model-family benchmark on the primary CUM+DYN representation. All models use the same landmark cohorts and athlete-disjoint test folds. XGBoost values are displayed to the precision exported by the frozen benchmark rerun.}
\label{fig:models}
\end{figure}

Figure~\ref{fig:models} and Table~\ref{tab:models} show that model ordering varies by landmark. TabPFN reaches its largest ROC-AUC at 30 min (0.692) and remains higher than logistic regression at 40--60 min. Paired athlete-cluster intervals for TabPFN minus logistic regression exclude zero for ROC-AUC at 40, 50, and 60 min and for PR-AUC at 50 and 60 min. In contrast, no paired TabPFN-versus-Random-Forest interval excludes zero; Random Forest is pointwise higher at 60 min. Against XGBoost, the clearest canonical contrast is at 50 min (TabPFN minus XGBoost: $\Delta$ROC-AUC 0.155, 95\% CI 0.055--0.253; $\Delta$PR-AUC 0.00336, 95\% CI 0.00046--0.00846). The result supports later-landmark improvement over logistic regression, not universal TabPFN dominance.

\begin{table}[t]
\centering
\caption{Canonical matched model benchmark. Values are ROC-AUC / PR-AUC.}
\label{tab:models}
\resizebox{\linewidth}{!}{%
\begin{tabular}{lcccccc}
\toprule
Model & 10 & 20 & 30 & 40 & 50 & 60\\
\midrule
Logistic regression & .499/.0107 & .557/.0150 & .607/.0129 & .428/.0088 & .402/.0079 & .367/.0080\\
Random Forest & .505/.0095 & .613/.0126 & .627/.0132 & .631/.0133 & .534/.0101 & .631/.0150\\
XGBoost & .418/.0080 & .550/.0110 & .622/.0154 & .631/.0132 & .444/.0086 & .586/.0131\\
TabPFN & .565/.0109 & .635/.0136 & .692/.0154 & .674/.0144 & .599/.0119 & .610/.0129\\
\bottomrule
\end{tabular}}
\end{table}

\subsection{Synthetic Augmentation: Conditional, Not Universal}
\begin{figure}[H]
\centering
\begin{subfigure}[t]{0.49\linewidth}
\centering
\includegraphics[width=\linewidth]{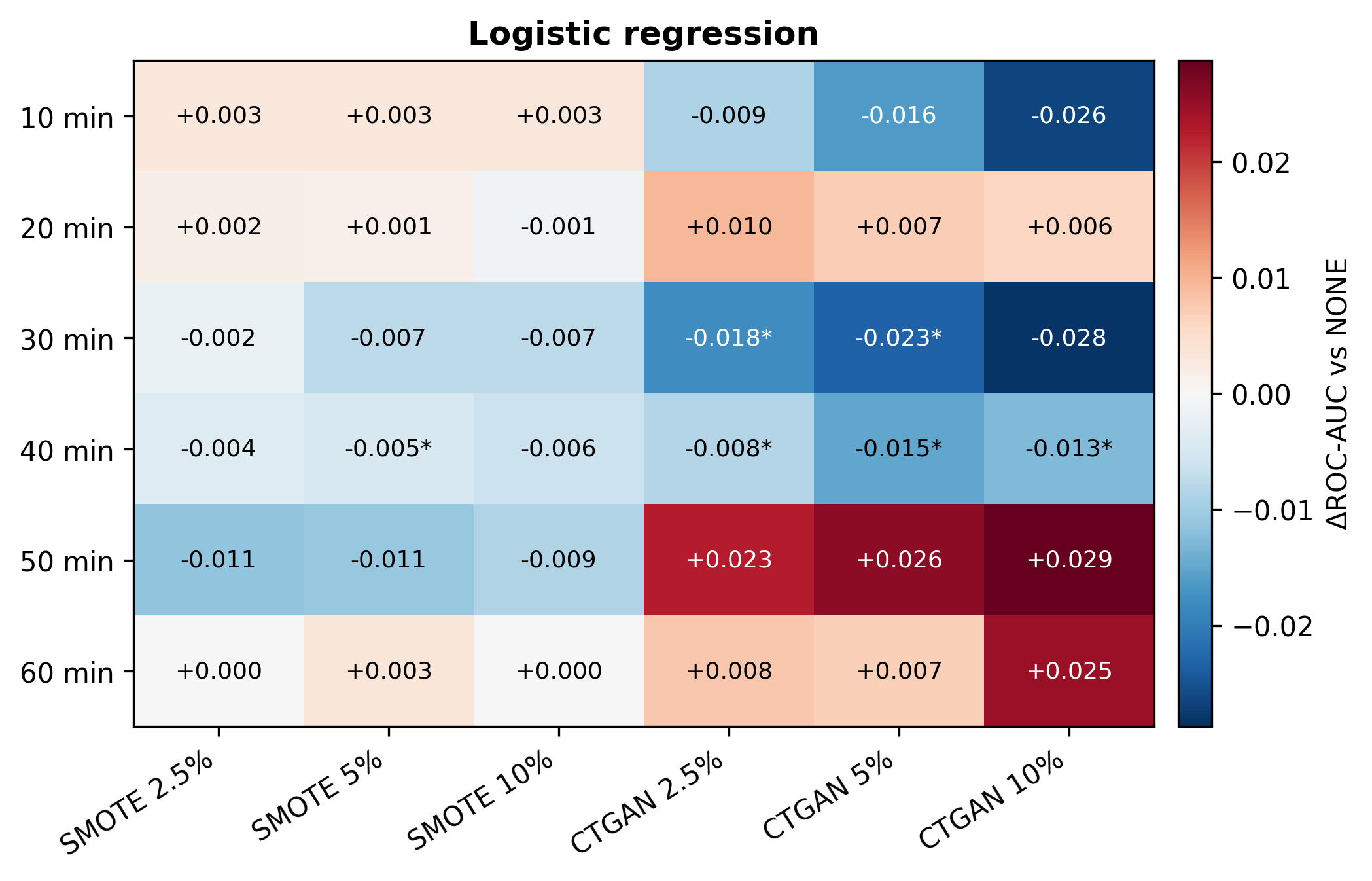}
\caption{Logistic regression}
\end{subfigure}\hfill
\begin{subfigure}[t]{0.49\linewidth}
\centering
\includegraphics[width=\linewidth]{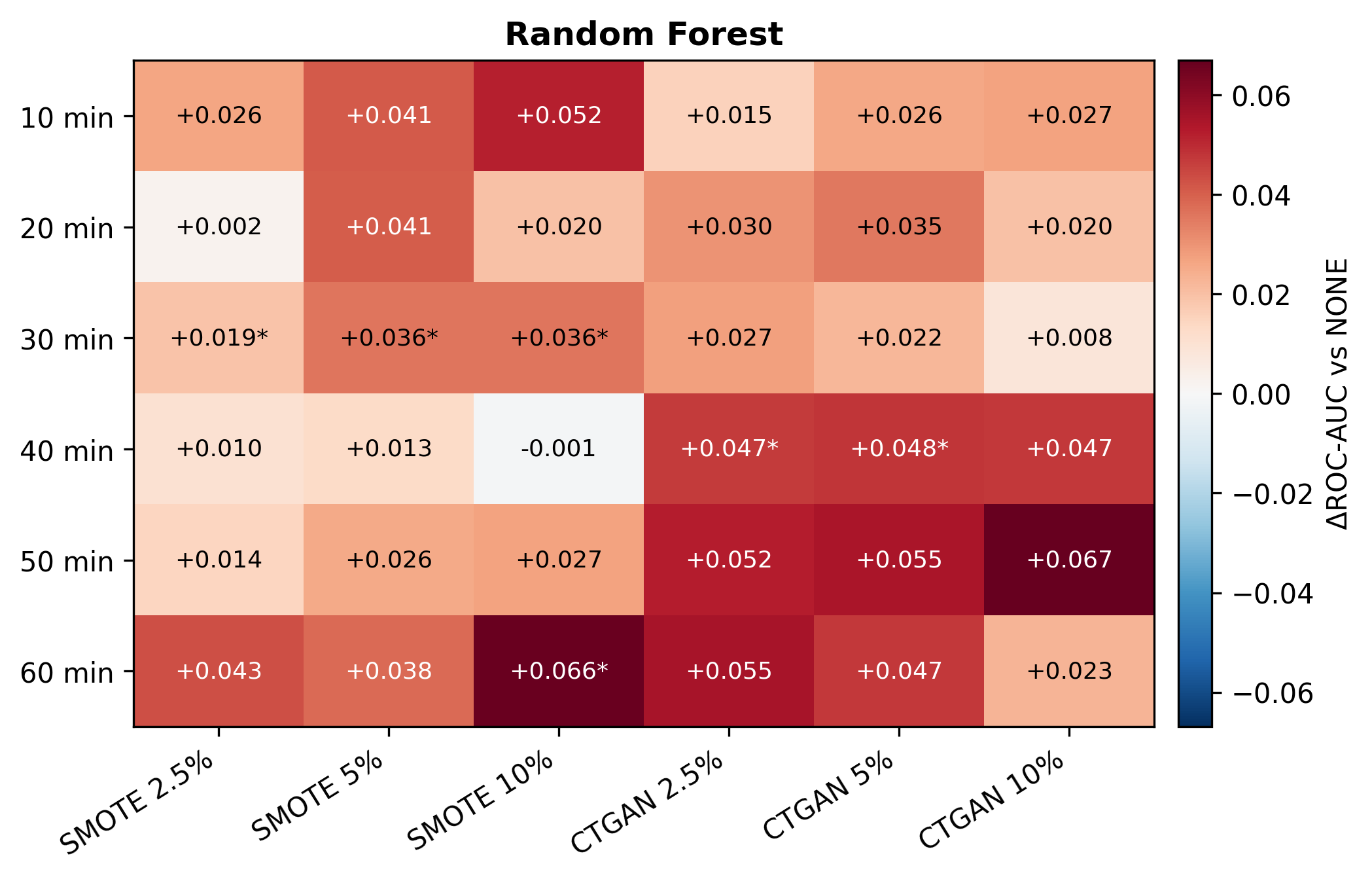}
\caption{Random Forest}
\end{subfigure}
\caption{Complete augmentation grid for mean $\Delta$ROC-AUC relative to NONE, averaged across five augmentation seeds. Columns show augmentation method and target synthetic ratio; rows show landmarks. Asterisks mark paired athlete-cluster bootstrap intervals that exclude zero. The figure visualizes all 72 prespecified model-by-landmark-by-method-by-ratio ROC contrasts rather than a selected subset.}
\label{fig:augmentation}
\end{figure}

The full augmentation grid does not support a universal benefit (Fig.~\ref{fig:augmentation}). For Random Forest at 30 min, SMOTE improves ROC-AUC across all three ratios, with paired intervals excluding zero: +0.0189 [0.0097, 0.0359], +0.0359 [0.0139, 0.0582], and +0.0360 [0.0185, 0.0653]. PR-AUC intervals also exclude zero for all three 30-min SMOTE ratios. At 40 min, Random Forest with CTGAN shows positive ROC-AUC contrasts at 2.5\% and 5\% (+0.0468 [0.0075, 0.1006] and +0.0479 [0.0053, 0.1232]); the 10\% ROC interval narrowly includes zero, although its PR interval excludes zero. Random Forest with 10\% SMOTE at 60 min shows a borderline positive ROC contrast (+0.0664 [0.00006, 0.0980]) while its PR interval includes zero.

Logistic regression behaves differently. CTGAN at 30 min produces negative ROC-AUC contrasts at 2.5\% and 5\% with intervals excluding zero, and all three CTGAN ratios at 40 min are negative with ROC intervals excluding zero; at 5\%, both ROC and PR intervals are negative. These patterns support learner- and landmark-dependent augmentation utility rather than a general class-imbalance solution.

\subsection{Augmentation Sensitivity and Synthetic-Data Audit}
Leave-one-positive-athlete-out summaries preserve the positive direction of Random-Forest+SMOTE at 30 min, Random-Forest+CTGAN at 40 min, and Random-Forest+SMOTE at 60 min across all five excluded positive athletes. Conversely, logistic-regression+CTGAN at 30 and 40 min remains directionally negative. Equal-athlete weighting agrees with the primary direction for 76.4\% of both ROC and PR augmentation contrasts, but some conditions reverse direction, including Random-Forest+CTGAN at 30 min. Augmentation effects are therefore partly estimand-dependent.

The CTGAN audit gives a complementary warning. Synthetic points are not near copies of real positives, but they are also not especially close to the observed positive distribution: the nearest-neighbour distance ratio is 1.49, median $|$SMD$|$ is 0.789, 49/63 features exceed 0.5, and median absolute correlation shift is 0.218. The absence of memorization is not evidence of fidelity. CTGAN is therefore treated as an experimental training intervention, not as a faithful generator of new injury cases.

\tightsection{Discussion}

The central methodological result is simple: \emph{minute-resolution predictors do not justify minute-resolution injury labels}. When the available outcome identifies only an injury-associated athlete-session, the supervised unit should remain the athlete-session. Fixed landmarks allow the representation to evolve with elapsed session time without fabricating an onset timestamp or turning every recorded minute into an independent labelled event.

Three empirical findings follow from that formulation. First, discrimination is landmark-dependent and non-monotonic. The primary logistic pattern rises through 30 min and then falls, and this shape persists under a fixed common cohort and 100 alternative negative-athlete allocations. More observed session time therefore does not imply monotonically better separation. This should not be interpreted as evidence that 30 min is a clinically optimal intervention point; exact biological onset is unknown.

Second, the apparent signal depends strongly on representation and estimand. PRE-containing configurations can yield materially larger point estimates than sensor-only CUM+DYN, but the associated missingness and timing ambiguity prevent operational pre-session claims. Equal-athlete weighting can also alter performance substantially because the primary session-pooled estimand gives more total influence to athletes who contribute more evaluable sessions. These are not technical footnotes: they change the population-level question being answered.

Third, greater modelling or synthetic-data complexity is not uniformly beneficial. TabPFN improves later-landmark discrimination relative to logistic regression, but does not consistently beat Random Forest. Similarly, augmentation produces localized gains for Random Forest yet little systematic improvement for logistic regression and several negative CTGAN effects. Synthetic augmentation therefore cannot be used as a rhetorical substitute for scarce independent positive athletes. It changes the training distribution; it does not increase the number of independent injury-positive people or medically adjudicated events.

\subsection{Scientific and Practical Scope}
The paper is intentionally not framed as a deployable injury-warning system. A practitioner can use the framework to ask a narrower and defensible question: given only monitoring information available by a fixed session-clock point, how separable are sessions that are associated with a same-day injury report from sessions that are not, when the evaluated athlete was never used for fitting? That question is meaningful for method development and study design even when the label is too coarse for minute-specific operational decisions.

The framework is also broader than football. Similar label-resolution mismatch can occur whenever sensors are sampled densely but outcomes are recorded at the shift, visit, day, bout, or session level. The core requirement is to align the supervised unit with the outcome resolution and to make clear which temporal information is observed versus which event timing is unknown.

\subsection{Limitations}
The decisive limitation is the effective positive sample: only 22 injury-associated sessions arise from five athletes. This constrains precision, stability, and external validity regardless of the 380,193 minute rows. Cluster bootstrap, leave-one-positive-athlete-out summaries, alternative folds, and synthetic augmentation expose sensitivity but cannot create missing biological replication.

Exact within-session injury-onset timestamps are unavailable, and injury outcomes are derived from athlete-submitted reports rather than medically adjudicated diagnoses. Consequently, an observation available by a landmark is session-clock-admissible but not guaranteed to precede biological injury onset. The study cannot localize injury onset, estimate prospective minute-specific risk, or evaluate an operational alerting policy.

All positive sessions occur in Team A; Team B contains no positive outcomes and cannot provide injury-positive external discrimination validation. Generalization across teams, leagues, sexes, age groups, or competitive levels is untested. The 22 positive dates should also not be interpreted as 22 confirmed independent injury events: repeated dates within an athlete may reflect distinct injuries, persistence, or repeated reporting.

PRE variables exhibit substantial missingness and imperfectly verified timing. Bootstrap intervals resample pooled out-of-fold predictions rather than refitting the full pipeline inside each bootstrap replicate. Finally, the augmentation experiment contains 72 exploratory contrasts with no prespecified multiplicity-adjusted confirmatory testing procedure. We therefore emphasize complete effect patterns, interval exclusion of zero, and robustness across estimands rather than isolated ``significant'' findings.

\tightsection{Conclusion}
We present a unit-aligned landmark framework for minute-resolution football monitoring when injury supervision exists only at athlete-session level. By constructing one representation per athlete-session at fixed landmarks, training only on other athletes, and keeping the target at its actual temporal resolution, the analysis preserves useful within-session information without inventing minute-specific supervision.

Within this cohort, discrimination depends on elapsed landmark, feature representation, learner, and weighting estimand. TabPFN improves later-landmark discrimination relative to logistic regression but does not consistently dominate Random Forest. Synthetic augmentation provides condition-specific benefits rather than a universal remedy, and CTGAN fidelity is limited despite no near-duplicate memorization signal. The principal remaining bottleneck is not algorithmic complexity but the scarcity and temporal coarseness of positive injury supervision. Larger independent cohorts with precise, medically adjudicated onset information are required before prospective injury-risk or deployment claims are justified.

\section*{Code and Reproducibility}
The public analysis repository contains the six-stage preprocessing, landmark construction, benchmark, TabPFN, and augmentation workflow used for the study. Raw SoccerMon data are not redistributed. Public code: \url{https://github.com/vaggoulas149/minute-level-injury-risk-monitoring}.

\end{document}